\pdfoutput=1  
\documentclass[letterpaper, 10 pt, conference]{ieeeconf}  
\IEEEoverridecommandlockouts                              
\usepackage{url}

\usepackage{gensymb}
\usepackage[utf8]{inputenc}
\usepackage{multirow}
\usepackage{chngcntr}
\counterwithin{table}{section}

\usepackage{graphicx} 
\usepackage{subfigure}
\graphicspath{ {./figures/} }

\newcommand{\SubfigImage}[3]{\subfigure[#3\label{fig:#2}]{\includegraphics[width=0.48\linewidth]{figures/#1}}}

\newcommand{\Figure}[3]{\begin{figure}[htbp]
  \centering
  #3
  \caption{#2}
  \label{fig:#1} 
\end{figure}}

\newcommand{\Figref}[1]{Fig. \ref{fig:#1}}
\newcommand{\figref}[1]{fig. \ref{fig:#1}}

\usepackage[usenames,dvipsnames]{xcolor}
\newcommand*{\REVISIONNEW}{}  %

\newcommand{\revision}[3]{%
\ifdefined\REVISIONCOMMENTS{ \color{cyan}{#1}\color{black} }\else\fi%
\ifdefined\REVISIONORIGINAL{ \color{gray}{#2}\color{black} }\else\fi%
\ifdefined\REVISIONHIGHLIGHT \color{red} \else \fi%
\ifdefined\REVISIONNEW {#3} \else \fi%
\color{black}%
}

\title{\LARGE \bf
Google Scanned Objects: \\
A High-Quality Dataset of 3D Scanned Household Items
}

\author{
Laura Downs$^{1}$, Anthony Francis$^{1}$, Nate Koenig$^{3}$, Brandon Kinman$^{1}$, Ryan Hickman$^{1}$,\\
Krista Reymann$^{1}$, Thomas B. McHugh$^{2}$,
and Vincent Vanhoucke$^{1}$
\thanks{$^{1}$
        Laura Downs,
        Anthony Francis,
        Brandon Kinman,
        Ryan Hickman,
        Krista Reymann,
        and Vincent Vanhoucke
        are with Robotics at Google, Mountain View, CA 94043, USA 
    (email:
        ldowns@google.com,
        centaur@google.com,
        bkinman@google.com,
        rhickman@google.com,
        reymann@google.com,
        vanhoucke@google.com)}%
\thanks{$^{2}$
        Thomas B. McHugh
        is with Northwestern University, Evanston, IL 60208, USA
    (email: mchugh@u.northwestern.edu)}%
\thanks{$^{3}$
        Nate Koenig
        is with Open Robotics, Mountain View, CA 94041, USA
    (email: nate@openrobotics.com)}%
}

\begin{document}

\maketitle
\thispagestyle{empty}
\pagestyle{empty}

\begin{abstract}

Interactive 3D simulations have enabled breakthroughs in robotics and computer vision, but simulating the broad diversity of environments needed for deep learning requires large corpora of photo-realistic 3D object models. To address this need, we present Google Scanned Objects, an open-source collection of over one thousand 3D-scanned household items released under a Creative Commons license; these models are preprocessed for use in Ignition Gazebo and the Bullet simulation platforms, but are easily adaptable to other simulators. We describe our object scanning and curation pipeline, then provide statistics about the contents of the dataset and its usage. We hope that the diversity, quality, and flexibility of Google Scanned Objects will lead to advances in interactive simulation, synthetic perception, and robotic learning.

\end{abstract}


\section{Introduction}

Deep learning has enabled many recent advances in computer vision and robotics, but
training deep models requires diverse inputs in order to generalize to new scenarios
\cite{DBLP:journals/corr/abs-2011-03395}.
Computer vision has used web scraping to gather datasets with millions of items, including
ImageNet \cite{DBLP:journals/corr/RussakovskyDSKSMHKKBBF14},
Open Images \cite{DBLP:journals/corr/abs-1811-00982},
Youtube-8M \cite{DBLP:journals/corr/Abu-El-HaijaKLN16},
and COCO \cite{DBLP:journals/corr/LinMBHPRDZ14};
however, labeling these datasets is labor-intensive,
labeling errors can distort the perception of progress \cite{DBLP:journals/corr/abs-2006-07159},
and this strategy does not readily generalize to 3D or real-world robotic data.
Unlike images, the web does not contain a large population of high-quality 3D scenes,
real-world data collection is challenging as robots are expensive and dangerous,
and human labelers cannot extract 3D geometric properties from images.

Simulation of robots and environments, using tools such as
Gazebo \cite{Koenig2004},
Bullet \cite{coumans2021},
MuJoCo \cite{todorov2012mujoco},
and Unity \cite{DBLP:journals/corr/abs-1809-02627},
can mitigate many of these limitations,
as simulated environments can be varied safely,
and semantic labels can be easily derived from the simulation state.
However, simulation is always an approximation to reality:
handcrafted models built from polygons and primitives correspond poorly to real objects.
Even if a scene is built directly from a 3D scan of a real environment,
the discrete objects in that scan will act like fixed background scenery
and will not respond to inputs the way that real-world objects would. 
A key problem, then, is providing a library of high-quality models of 3D objects
which can be incorporated into physical and visual simulations
to provide the required variety for deep learning.

To address this issue, we present 
the Google Scanned Objects (GSO) dataset,\footnote{Dataset available at
https://goo.gle/scanned-objects}
a curated collection of over 1000 3D scanned common household items
for use in the 
Ignition Gazebo \cite{Koenig2004} and
Bullet \cite{coumans2021}
simulators, as well as other tools that can read the SDF model format. 
In this letter, we describe our pipeline for 
object collection and curation,
scalable, high-quality 3D scanning,
scan quality assurance and publishing.
In addition, we present breakdowns of
the statistics of the objects in the dataset
and the usage of the dataset in published research.

Our contributions include
(a) the Google Scanned Objects dataset,
(b) the design of our 3D scanning pipeline,
(c) the design of our 3D scan curation and publication process,
and (d) a review of the impact of this dataset on research.

\begin{figure}[t]
\centering
\includegraphics[width=0.48\textwidth]{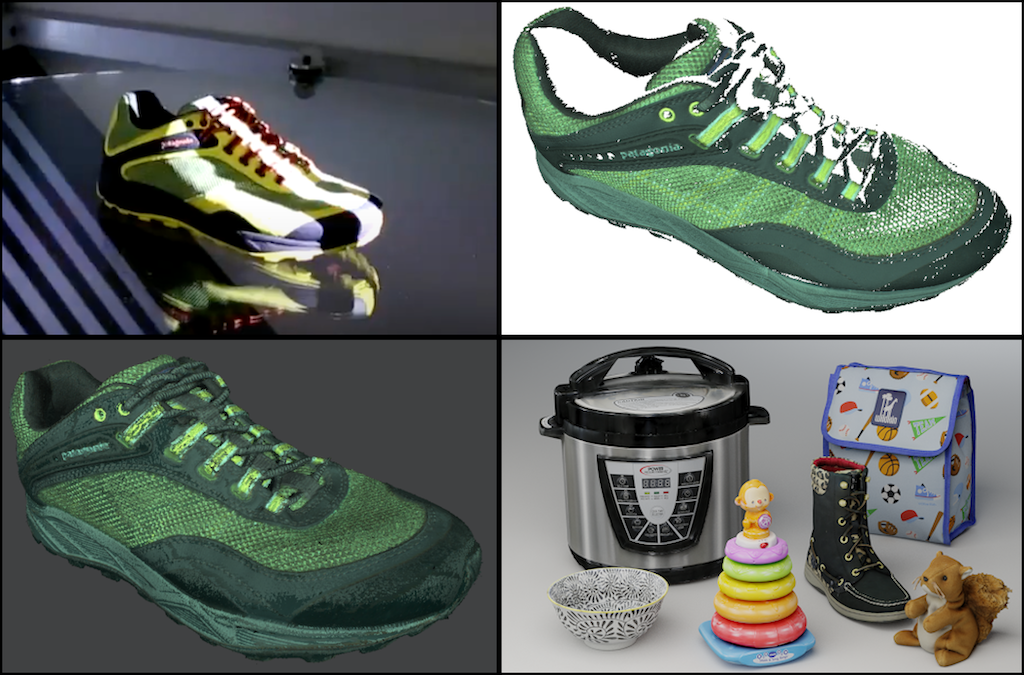}
\caption{
Custom 3D scanning hardware enabled fast capture of raw meshes,
which our scanning pipeline aligned using a calibration process
followed by QA curation of high-quality models for inclusion in the dataset.
}\label{fig:scanning-examples}
\end{figure}

\begin{figure*}[htbp]
  \centering
  \subfigure[Scanning Pipeline\label{fig:scanning-pipeline}]{
  \includegraphics[width=0.35\linewidth]{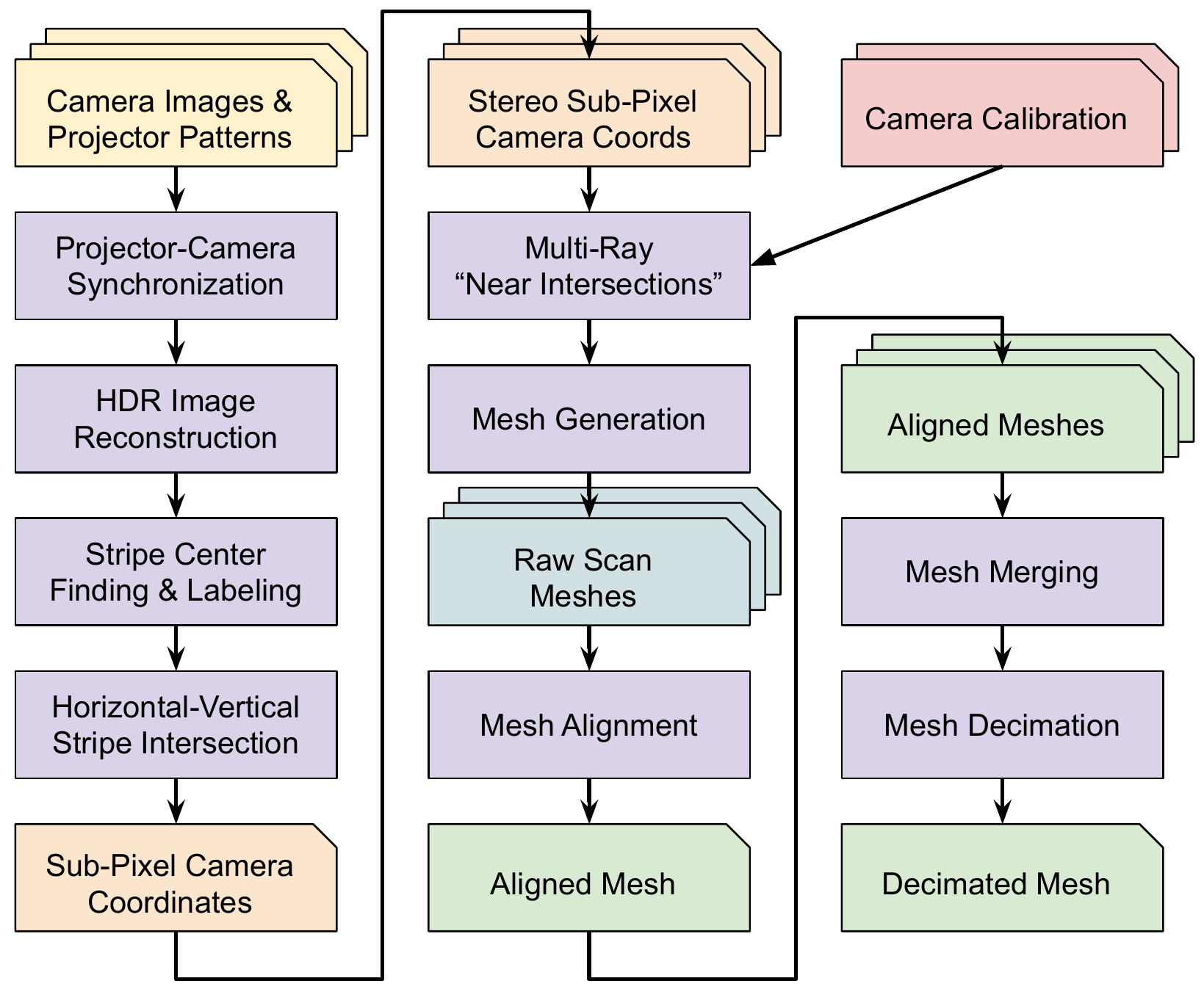}}
  \subfigure[Scanner Hardware\label{fig:scanner-hardware}]{
  \includegraphics[width=0.22\linewidth]{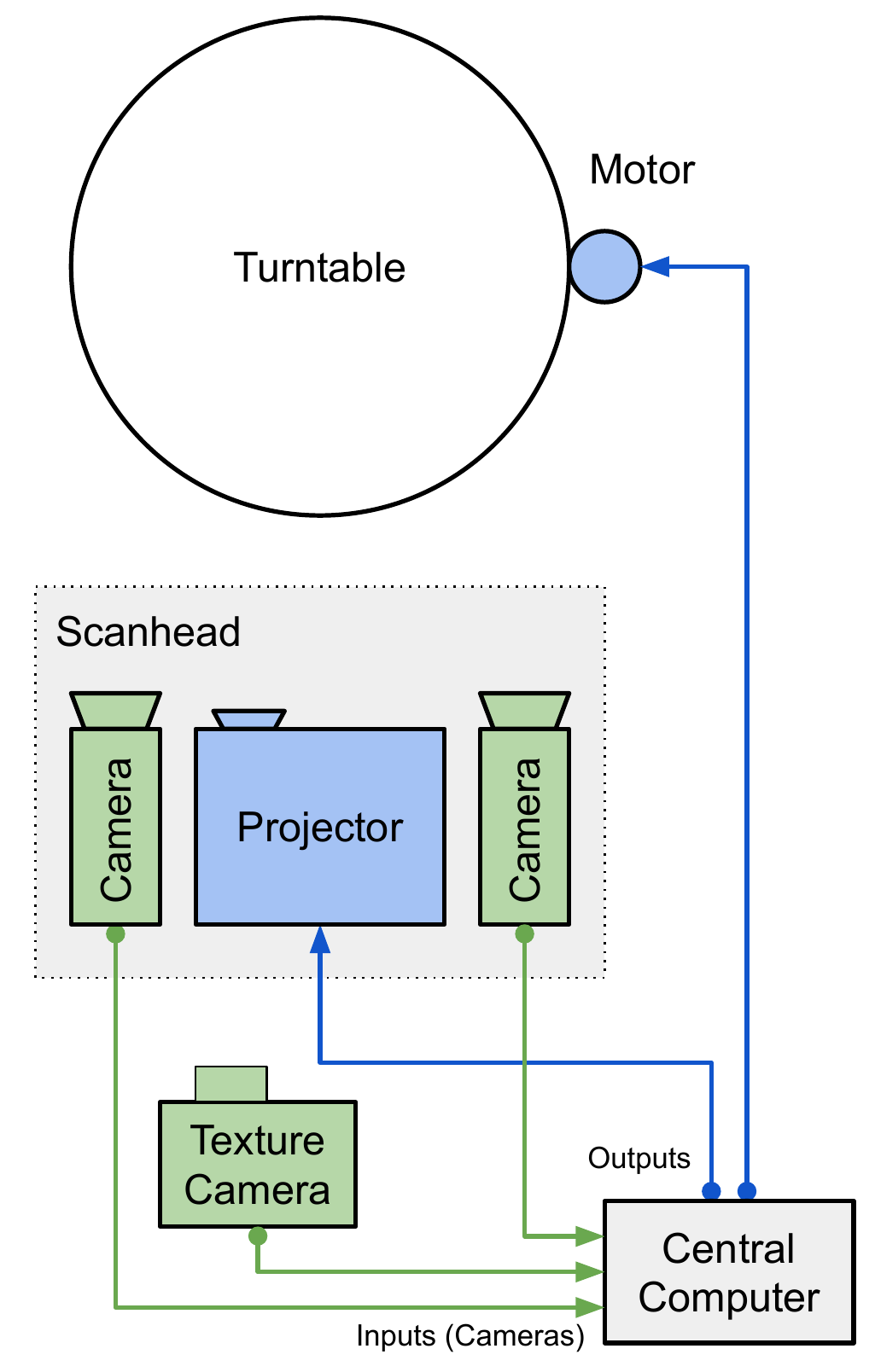}\hfill
  \includegraphics[width=0.34\linewidth]{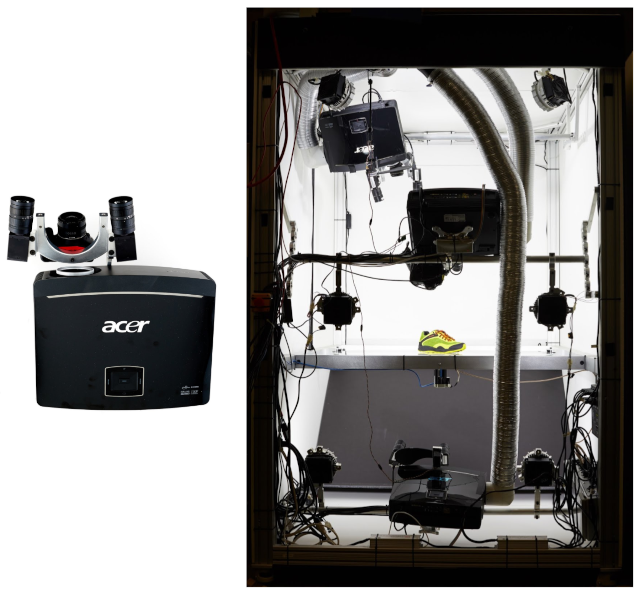}}

  \caption{Google Scanned Objects'
  custom 3D scanning pipeline (a) first reconstructed sub-pixel camera coordinates for each camera
  before merging calibrated stereo pairs into raw scan meshes for further processing.
  A custom lighting-controlled enclosure (b)
  included two machine vision cameras for stereo shape detection,
  a DSLR camera for high-quality HDR color frame extraction,
  and a computer-controlled projector for gray codes and stripe patterns.}
  \label{fig:scanning-overview} 
\end{figure*}

\revision{}{}{
\begin{table}[tbp]
\centering
\caption{Selected 3D Model Databases Well-Suited for Simulation.}
{\footnotesize\begin{tabular}{|l|c|c|c|c|c|}\hline
\textbf{Dataset} & \textbf{Year} & \textbf{Purpose} & \textbf{Cat.} & \textbf{Obj.} & \textbf{Img.}  \\ \hline
\multicolumn{6}{|l|}{\bf 3D CAD Models} \\ \hline
PhotoShape \cite{photoshape2018} & 2020 & 3D Graphics & 1 & 11,000 & 40,927  \\
3D-Future \cite{fu20203dfuture} & 2020 & Shape Retr. & 1 & 16,563 & 20,240  \\
ABO \cite{collins2021abo} & 2021 & Shape Retr. & 98 & 147,702 & 398,212  \\ \hline
\multicolumn{6}{|l|}{\bf 3D Scanned Models} \\ \hline
KIT \cite{kasper2012kit} & 2012 & Object Reco. & N/A & 145 & 196188  \\
BigBIRD \cite{singh2014bigbird} & 2014 & Object Reco. & N/A & 125 & 75,000  \\
YCB \cite{calli2015benchmarking} & 2015 & Object Reco. & 5 & 77 & 92,400  \\
\textbf{GSO (ours)} & \bf 2021 & \bf Simulation & \bf 17 & \bf 1,030 & \bf N/A  \\ \hline
\end{tabular}}
\label{table:database-comparison}
\end{table}
}

\section{Related Work}
Many simulators are available for robotics applications \cite{9386154},
and many learning systems have used simulation to train models to deploy on robots.
While early work used static environments in simulators similar to Bullet \cite{coumans2021},
more recent work has injected 3D objects into the environment to enable training interactive navigation,
such as the Interactive Gibson benchmark \cite{8954627} which uses GSO.

3D objects for simulation have been traditionally created through
manual 3D modeling \cite{DBLP:journals/corr/abs-2106-14405}, 
object scanning \cite{levoy2000digital}, 
conversion of CAD models \cite{SolidWorksToUrdf},  
or combinations of these techniques \cite{FacebookAndMatterport}.
\revision{[E1, R1.1, R2.1, R3.1]}{%
Some manufacturers provide 3D models of their products,
such as Steelcase \cite{SteelcaseCAD},
Siemens \cite{SiemensCAD},
and IKEA \cite{IKEACAD}.
These models have proved useful in virtual environment research \cite{bukowski1997interactive}
as they often feature articulated 3D geometry difficult to capture via scanning.
New neural rendering techniques, such as 
neural radiance fields \cite{DBLP:journals/corr/abs-2003-08934}, 
have also shown promising results for 3D object scanning.
}{%
For example, \cite{DBLP:conf/si3d/FunkhouserST92} used several techniques to create furniture models for a virtual environment building walkthrough.
New neural rendering techniques, such as 
neural radiance fields \cite{DBLP:journals/corr/abs-2003-08934}, 
have also shown promising results for 3D object scanning.

3D objects have been collected into datasets for many purposes:
\cite{calli2015benchmarking} describes 33 such datasets, though not all remain extant (for example, the Amazon Picking Challenge dataset \cite{correll2016analysis} is no longer readily available).
These datasets serve needs such as shape and object retrieval \cite{zhang2005retrieving},
object recognition and computer vision \cite{singh2014bigbird},
grasping and manipulation \cite{calli2015benchmarking},
and even research into prosthetics and rehabilitation \cite{kapadia2012toronto}.
Regardless of its original purpose, a dataset can be useful in a simulation system if it is currently available and contains objects in watertight meshes, ideally with high visual fidelity. Datasets that provide only scenes \cite{Dai_2017_CVPR}, objects in scenes \cite{meger2013ubc, janoch2013category, xiang2016objectnet3d, objectron2021, reizenstein21co3d} or object point clouds \cite{lai2011large} are less comparable to our dataset.
 
Some manufacturers provide 3D models of their products that are usable in simulation;
these can sometimes describe articulated 3D geometry, mechanical properties, and other features difficult to capture via scanning.
Manufacturers such as 
Steelcase \cite{SteelcaseCAD} and
Siemens \cite{SiemensCAD}
provide CAD shape models, while  
IKEA \cite{IKEACAD}
provides models which often include rich textures.
The IKEA dataset has been used for both shape analysis \cite{lim2013parsing} and as a foundation for further dataset creation, such as the IKEA ASM furniture assembly dataset \cite{ben2021ikea}.
ModelBank also provides 1200 high-quality licensable models \cite{ModelBank}.

3D object datasets suitable for simulation research include those composed of 3D CAD models \cite{shilane2004princeton, zhang2005retrieving, tatsuma2012large, xiang_wacv14,  shapenet2015, mo2019partnet, photoshape2018, fu20203dfuture, collins2021abo} and 3D scanned models \cite{kasper2012kit, Lian2011, li2012shrec, Pickup2014, gao2015shrec, singh2014bigbird, calli2015benchmarking, knapitsch2017tanks, pix3d} though many combine types \cite{li2012shrec}. 
Datasets most comparable to GSO include freely available, watertight, textured models with high visual fidelity usable in simulation. Scanned datasets include YCB \cite{calli2015benchmarking} with 77 objects in 5 categories, BigBird \cite{singh2014bigbird} with 125 objects, and KIT \cite{kasper2012kit} with 145 objects; however these datasets generally do not include object bases. CAD datasets include PhotoShape \cite{photoshape2018} with 11,000 CAD chairs, 3D-FUTURE \cite{fu20203dfuture} with 16K CAD chairs, and most notably the recently released Amazon Berkeley Objects dataset with 8K CAD objects in almost 100 categories \cite{collins2021abo}.
Table \ref{table:database-comparison} quantitatively compares these datasets on year of release as well as number of categories, objects, and reference images.}

Some 3D datasets have proved useful in supporting grasping tasks.
ShapeNet collects 51,300 3D CAD models into 55 categories
within the WordNet taxonomy \cite{shapenet2015}.
ShapeNet is designed for vision and does not contain mass or friction, 
but GraspGAN \cite{8460875} trained a grasping system
using photo-realistic renders based on both ShapeNet and procedural models.
For GraspGAN, procedural models worked better than ShapeNet's 3D CAD models
(though ShapeNet has a stated goal of adding 3D scanned content).
The YCB (Yale-CMU-Berkeley) Dataset \cite{calli2015benchmarking}
is 77 real-world graspable objects and
parallel model database of textured meshes
to facilitate sim-to-real domain adaptation.
\revision{[E1, R1.1, R2.1, R3.1]}{%
These meshes are generally lower fidelity than GSO.
}{
The KIT object models database \cite{kasper2012kit} uses a setup similar to GSO to capture 145 scanned objects with calibrated images and grasp data for commercial manipulators; however, this scanner does not capture object bases.
}

Several other 3D object datasets have been created for computer vision.
Objectron \cite{objectron2021} 
consists of object-centered videos and augmented reality 3D pose information,
used to create machine learning pipelines for 3D object bounding box detection.
The Berkeley 3-D Object Dataset (B3DO) 
\cite{janoch2013category} 
collects over 800 3D scanned models in over 50 categories,
scanned via the Kinect 3D sensor and
annotated via the Amazon Mechanical Turk (AMT) \cite{paolacci2010running}.
ScanNet is a collection of over 1500 3D scanned room-sized scenes collected
via a similar Kinect-AMT pipeline, augmented with dense 3D reconstructions,
and PartNet \cite{mo2019partnet} consists of over 25K 3D CAD models
broken down into almost 575K constituent parts;
both have been used for semantic segmentation.
ObjectNet3D \cite{xiang2016objectnet3d} 
consists of 44K 3D shapes in 100 categories, aligned with  90K images
annotated with 3D poses, and has been used for object detection.
\revision{[E1, R1.1, R2.1, R3.1, R3.2]}{%
}{%
BigBIRD, the (Big) Berkeley Instance Recognition Dataset \cite{singh2014bigbird} contains 125 3D meshes, each with 600 calibrated RGB-D images and segmented point clouds to facilitate object recognition, captured with a setup similar to GSO; like KIT, however, these meshes lack views of object bases. 3D-Future \cite{fu20203dfuture} uses 16K CAD textured chairs to generate 20K realistic images corresponding to 5K scenes, designed to enable precise comparison of images to models for benchmarking. The Amazon Berkeley Objects \cite{collins2021abo} dataset contains 8K high-quality textured CAD objects in 98 categories, along with reference images; this dataset, along with GSO, was used to benchmark a differentiable stereopsis shape reconstruction approach \cite{goel2021differentiable}.}

\section{Creation of the Dataset}

The GSO dataset began as an effort by Google's Cloud Robotics initiative in 2011
to enable robots to recognize and grasp objects in their environments using
high-fidelity 3D models of common household objects. 
However, 3D models have many uses beyond object recognition and robotic grasping, including scene
construction for physical simulations and 3D object visualization for end-user applications.
Therefore, Google Research initiated a project to bring 3D experiences to Google at scale,
collecting a large number of 3D scans of household objects
at a cost less than that of traditional commercial-grade product photography.

This was an end-to-end effort, including object acquisition, novel 3D scanning hardware,
efficient 3D scanning software, fast 3D rendering for QA, web and mobile viewers, and human-computer
interaction studies to create effective experiences for interacting with 3D objects. Following
data collection, we then built a pipeline to make this data available in a variety of formats. In the
following sections, we discuss the 3D scanning innovations necessary to collect scans of household items,
followed by the pipeline necessary to make them available to simulation systems.

\Figure{calibration-process}{
    Scanning as a calibration process.
    (a) Calibration patterns enabled our 2D pipeline to accurately align cameras.
    (b) A computer-controlled projector created similar patterns for 3D scanned objects.
    (c) Appropriate patterns enable detecting locations with sub-pixel accuracy.
    (d) A large suite of patterns enabled extracting the full 3D shape of scanned objects.
    }{%
\SubfigImage{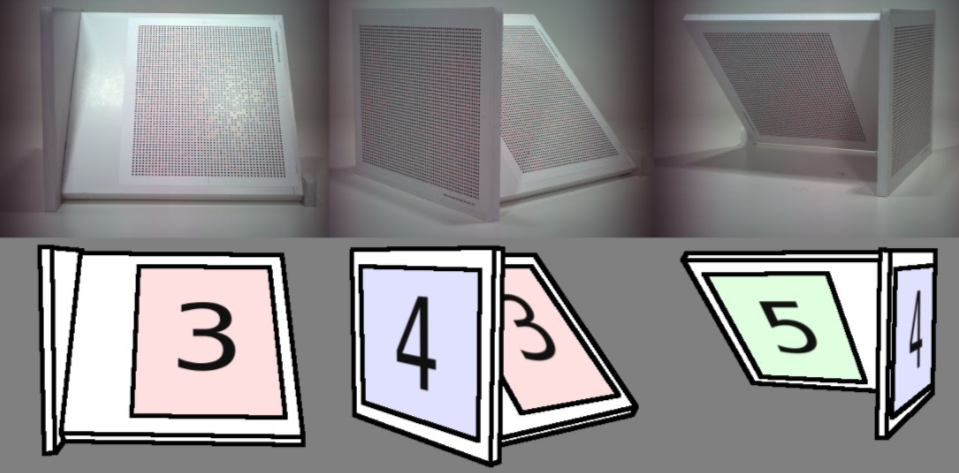}{scanner-calibration}{Calibration patterns.}
\hfill
\SubfigImage{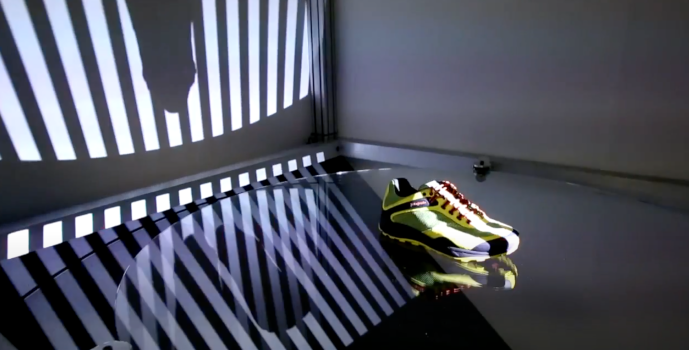}{projected-patterns}{Projected 3D patterns.}%
\\
\SubfigImage{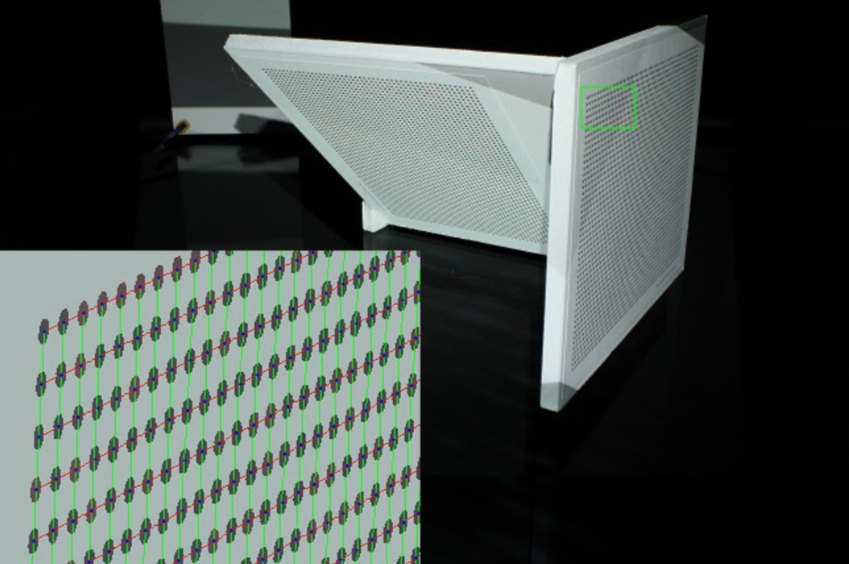}{pattern-detection}{Pattern detection.}
\hfill
\SubfigImage{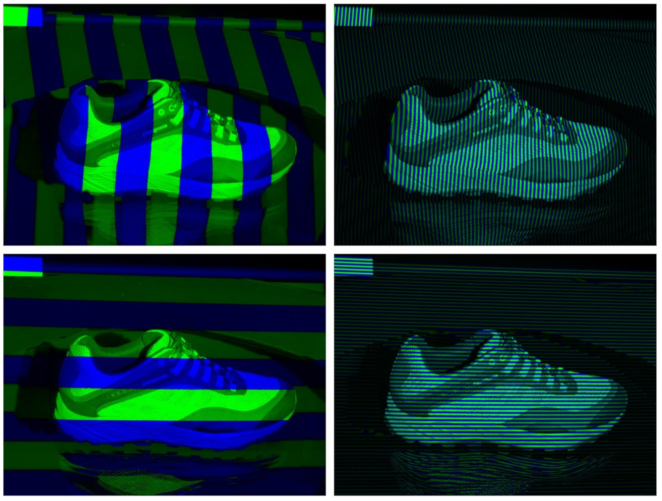}{hdr-vs-graycodes}{HDR vs gray-codes.}}

\Figure{good-vs-bad}{
    Scanned items required a QA pass.
    (a) Many items were captured as high-quality closed manifold meshes.
    (b) Optically uncooperative objects rarely yielded invalid meshes, but sometimes exhibited deformations.
    }{%
\SubfigImage{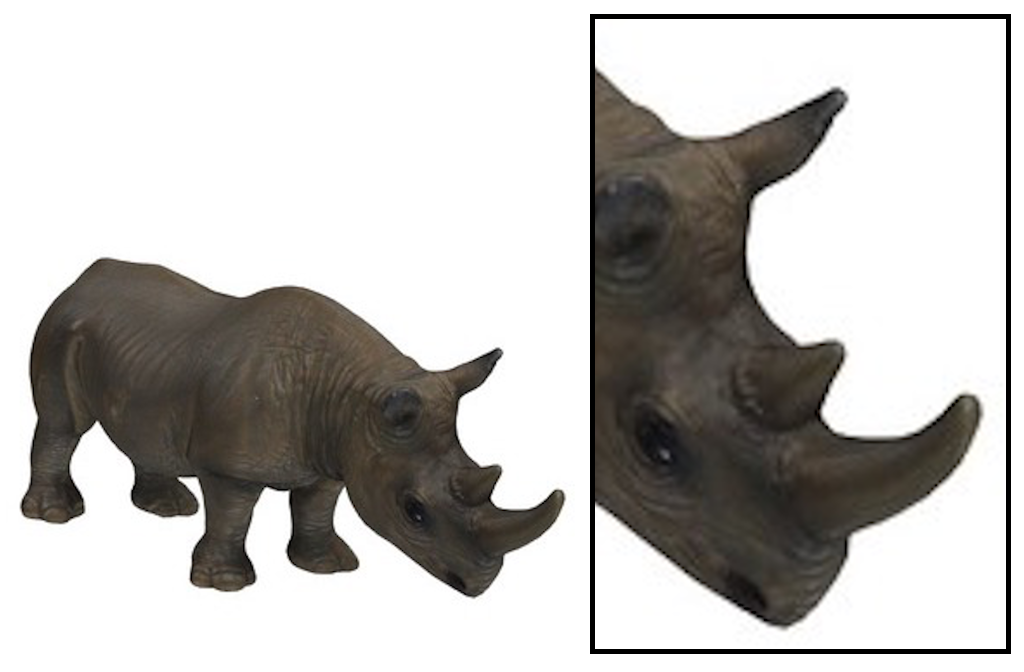}{good-scan}{Closed manifold mesh.}
\hfill
\SubfigImage{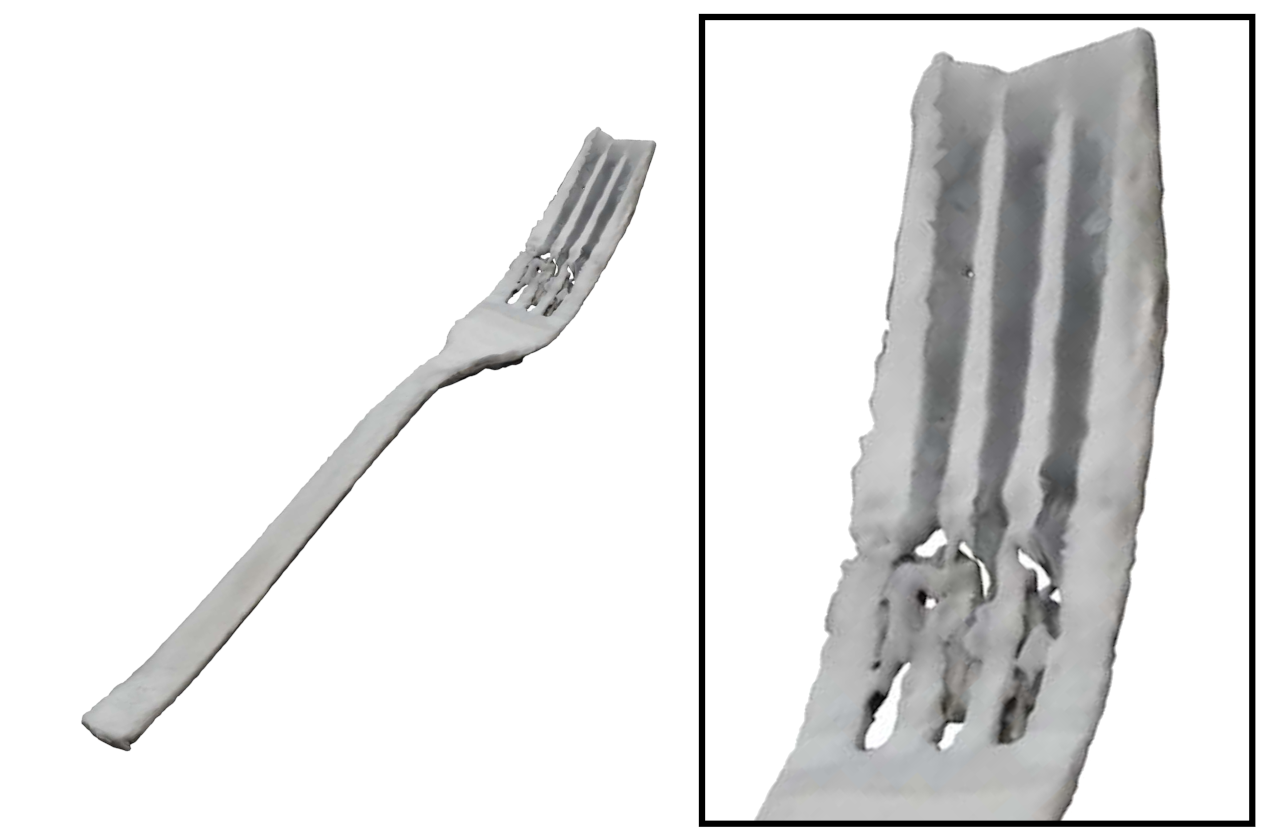}{bad-scan}{Deformations due to reflections.}}

\subsection{3D Scanning Pipeline}
Even when focusing only on household objects, 3D scanning presents distinctive challenges, including
efficient physical scanning setups, target lighting, camera reliability, scanner performance,
color matching, texture rendering, and dealing with 
optically uncooperative materials, such as
near-white, shiny, or transparent surfaces.
Dedicated 3D scanning hardware existed at the time but was labor-intensive and not cost-effective.
To scan at scale, we needed something that functioned less like a scientific
instrument and more like a microwave in its usability and its reliability.

Therefore, we designed our own dedicated scanning hardware and software (\Figref{scanning-overview}),
capable of scanning objects in 10 minutes and producing high-resolution models with 2M triangles.
The lighting-controlled physical enclosure (\Figref{scanner-hardware})
captures 3D geometry using structured light scanning
with two machine vision cameras and a projector, and captures textures with product-friendly
lighting with a separate DSLR high-resolution camera. By the end of our project's first year,
we were capturing over 400 scans per week, and over the course of the project, we captured
100K $360\degree$ photo swivels and 10K fully-3D scans of unique objects.

Our scanning software pipeline treats 3D scanning as a calibration problem (\Figref{scanning-pipeline}),
using libraries shared with our $360 \degree$ swivel scanning pipeline.
To stitch multiple 2D images into a smooth animated rotation,
the $360 \degree$ swivel pipeline scans 3D reference calibration patterns as if they are ordinary objects (\Figref{scanner-calibration}),
enabling camera geometry to be calibrated with sub-pixel accuracy (\Figref{pattern-detection}).

The 3D pipeline extends this process using a projector to generate over 200 patterns
(\Figref{projected-patterns}).
Building on an approach developed by Jens G\"uhring \cite{Guhring2001},
we use a stereo pair of machine vision
cameras to capture gray codes and phase patterns, projected in a range of
intensities, to build up HDR (High Dynamic Range) single-color pattern frames, augmented by a centrally
mounted DSLR which captures an aligned view of the color textures
(\Figref{hdr-vs-graycodes}).

Gray codes \cite{doran2007gray}, an indexing method robust to bit errors, are used
to identify projector pixels in camera images, but are not sufficient to guarantee
sub-pixel accuracy needed for texture mapping. Phase patterns (X and Y stripes)
identify exact projector lines to sub-pixel accuracy, but are not globally unique
without the gray codes for correspondence.

We use camera calibration and image reconstruction to build solid color HDR images
of the various patterns. These images, along with the original raw frames, form the
basis for parallel gray code indexing and stripe normalization pipelines. A convolution
operation enables us to find the true centers of the stripes, producing horizontal and
vertical code frames. To speed up the combination of horizontal and vertical frames into
joint $x,y$ coordinates, we took advantage of the fact that the original projected frames
have a known width and height in pixels.
Therefore, any given image can be viewed as having a known integer range
in one dimension and a to-be-determined floating point dimension in the other.
This permits a fast  $O(n^2 log(n))$ active interval sweep line algorithm which
requires effectively logarithmic per-pixel work for normal image sizes.

This algorithm produces \textit{pcoord} (projector coordinate) frames with sub-pixel
scale resolution, each representing a unique position on the projector.
Combined with a camera's location, each pcoord generates a camera ray.
Therefore, 2 or more pcoord pixels define a 3D point; the nearest point
to the 3D ``intersection'' has a fast closed-form solution for 2 rays. This
generates a depth map, which we use to generate a mesh by looking for valid
quad-pixel neighborhoods, outputting two mesh triangles per  ``quad'' found.
The result is a ``raw mesh'' from one camera view (Fig. \ref{fig:scanning-examples}, upper right).

The computer-controlled turntable enables the scanner to repeat the mesh generation process
at precise angles around an object, resulting in a set of raw meshes already in close alignment.
We use the Ceres solver \cite{ceres-solver} to implement a fast ICP solver
on all pairs in parallel, focusing on not breaking the good initial convergence.
This process optimizes 100K to 400K constraints per iteration and runs in less
than a minute even on the relatively large scans produced by our scanner. This
results in an aligned mesh (Fig. \ref{fig:scanning-examples}, lower left),
which the local scanner software then decimates,
assigns initial colors, and uploads to the cloud for further processing.

This pipeline enables the scanner to produce nearly complete raw scans directly on the hardware
as fast as it can physically
scan objects, allowing our remaining pipeline to focus on more intensive issues,
such as reconstruction of the inside of concave objects or human quality assurance.

\subsection{Simulation Model Conversion}
These original scanned models used 
protocol buffer metadata \cite{ProtocolBuffers},
very high resolution visuals,
and formats unsuitable for simulation.
Some physical properties, such as mass, were captured for many objects,
but surface properties, such as friction, were not represented in the metadata.

To enable these scanned models to be used in simulation systems, each was passed through
a pipeline which performed the following steps:
\begin{enumerate}
\item \textbf{Filter invalid objects.}
The dataset needed to be cleaned prior to processing. Duplicate items were removed,
along with test objects, calibration objects, and a few hand-modeled objects.
Models with invalid or missing meshes, textures or metadata were also excluded.
The remaining valid models went through a manual QA process prior to inclusion
(\Figref{good-vs-bad}).

\item \textbf{Assign object names.}
Human readable names were constructed automatically from text descriptions of the objects,
followed by manual QA to fix more mistakes.

\item \textbf{Validate object meshes.} 
Simulation requires a {\em closed manifold mesh},
a physically realizable volume described by a surface of polygonal facets
with an interior, exterior, no self-intersections, no holes, and non-zero thickness everywhere.
Furthermore, all geometry must be represented directly in the mesh, 
since bump maps, offset maps, and normal maps are not generally used in simulations.
Almost all objects in our dataset had closed manifold meshes, and the few
invalid meshes were excluded.

\item \textbf{Calculate physical properties.}
The original pipeline focused on visualization, not simulation, and did
not capture properties like object density or friction coefficients.
Starting with a clean mesh, we calculated the bounding box and volume.
While some objects had recorded mass, for the rest we estimated
mass from volume; similarly, we calculated center of mass and
moments of inertia assuming objects were solid and homogeneous.
When mass was not provided, we estimated a default constant density of 0.1 g/ml, which is a reasonable estimate for a hollow plastic object.
We left friction values unspecified, which should use the simulator default values.
While these estimates limit the use of these models for tasks which
require accurate physical parameters, they are sufficient to
enable 
\revision{}{the use of these models}{their use} in many simulation platforms.

\item \textbf{Construct collision volumes.}
We explored several techniques to generate collision volumes for our models,
including VHACD (Volumetric Hierarchical Approximate Convex Decomposition) \cite{5414068}
to generate convex collision volumes for use in simulation tools that require a convex decomposition.
However, we identified different use cases which would require more or less complex collision meshes,
and ultimately chose to use the visual mesh directly as the collision volumes.

\item \textbf{Reduce model size.}
Our scanning pipeline produces meshes and with high visual fidelity
but these needed to be simplified to reasonable resolutions for rendering
using edge collapse algorithms with quadratic error metrics \cite{garland1997surface}.
The smallest of the simplified meshes provided by the original scanning pipeline
down-sampled the original million-vertex meshes to tens of thousands of vertices
and were generally suitable for physical simulations.
We used Blender \cite{BlenderDecimateModifier} to further down-sample meshes and textures for internal projects.

\item \textbf{Create SDF models.}
These scans did not include linkage or joint information,
so a simple model in SDF format was created for each object
with one link to one OBJ visual mesh and one collision mesh.

\item \textbf{Create thumbnail images.}
We mined thumbnail images from the preview images of the original dataset.

\item \textbf{Package the model.}
For use in simulators or in public interfaces with searchable metadata,
each model had to be self-contained with a metadata file and all related assets.
Associated file inputs such as visual meshes, collision meshes, and textures
were organized by a relative path from the model directory.
\end{enumerate}

The output of this pipeline was 
a simulation model in an appropriate format
with a name, mass, friction, inertia, and collision information, along with
searchable metadata in a public interface compatible with Ignition Fuel.

\subsection{Ignition Fuel}

To host the models, we chose Ignition Fuel, an open-source hosting environment that implements an evolution of Gazebo’s model hosting strategy. Historically, Gazebo used models hosted on version-controlled repositories such as Git \cite{chacon2014pro}. This requires minimal development and maintenance, but does not scale to large numbers of models. Code repositories are not designed for large binary datasets or discoverability, making storing and finding models cumbersome.

To address these issues, Open Robotics developed Ignition Fuel, which encapsulates each model within its own
Git repository, hosted on a server with a scalable filesystem. This approach provides the benefits of model version control while limiting repository size. A front-end web application provides a user-friendly interface to the server with features including account management, model upload and download, 3D visualization of a model, and search.
Each model released to Ignition Fuel includes an OBJ visual mesh, MTL material file, PNG diffuse texture map, SDF model description, text protobuf metadata, and thumbnail images. 

To host GSO, Open Robotics collaborated with Google to expand Ignition Fuel's metadata, to scale to larger sets of models, and to improve model discoverability and accessibility through ElasticSearch. 
Ignition Fuel model metadata includes model name, description, and authors; for GSO, we expanded this to include SKUs, manufacturer part numbers, brand, tool-chain compatibility, copyright and license information, tags, version, annotations and categories. This additional metadata supports improved search and vetting of models, and is captured in a protobuf text file alongside other model assets.

In addition, Ignition Fuel Tools is a command line interface (CLI) which communicates with the Ignition Fuel server through its REST API, presenting a more user-friendly interface than general purpose tools such as curl. For GSO, we expanded the CLI to support bulk upload and downloads via collections - a logical grouping of models such as the GSO dataset.

\begin{table}[tbp]
\centering
\caption{Model complexity by category.}
{\footnotesize\begin{tabular}{|l|c|c|c|}\hline
Category                  & Models & Vertices: avg. $\pm\ \sigma$ & min / max \\ \hline
Action Figures            &         17       & 10284  $\pm$  2643 & 5413 / 15621 \\
Bag                       &         28       & 9450 $\pm$ 3951 & 5511 / 21922  \\
Board Games               &         17       & 2137 $\pm$ 2164 & 513 / 9861    \\
Bottles/Cans/Cups         &         53       & 4984 $\pm$ 1860 & 1035 / 13638  \\
Camera                    &         1        & 4785 &      \\
Car Seat                  &         1        & 16416 &    \\
Consumer Goods            &         248      & 3465 $\pm$ 3874 & 577 / 42890   \\
Hat                       &         2        & 19313 $\pm$ 7915 & 11398 / 27229\\
Headphones                &         4        & 5733 $\pm$ 5314 & 1995 / 14819  \\
Keyboard                  &         4        & 10682 $\pm$ 3709 & 4280 / 13251 \\
Block toys                     &         10       & 13477 $\pm$ 13072 & 942 / 41036 \\
Media Cases               &         21       & 2557 $\pm$ 2033 & 1166 / 10293  \\
Mouse                     &         4        & 13386 $\pm$ 2888 & 9686 / 16620 \\
Shoe                      &         254      & 13091 $\pm$ 4931 & 3558 / 32555  \\
Stuffed Toys              &         3        & 31062 $\pm$ 12617 & 18237 / 48220\\
Other toys                &         147      & 7684 $\pm$ 4829 & 1419 / 36138   \\
Uncategorized             &         216      & 8216 $\pm$ 8669 & 828 / 50561    \\ \hline
\end{tabular}}
\label{table:model-stats}
\end{table}

\Figure{object-details}{\revision{[E1, R1.1, R2.1, R3.1]}{}{Because our scanner uses calibrated projected patterns rather than raw depth data, the GSO process captures fine surface details (a) and complex topology (b) in scanned models.}}{
\SubfigImage{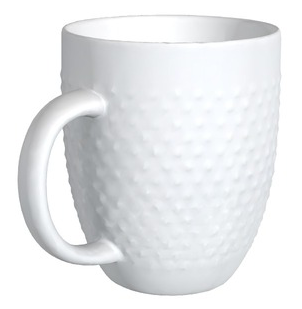}{white-cup}{Fine surface details.}
\hfill
\SubfigImage{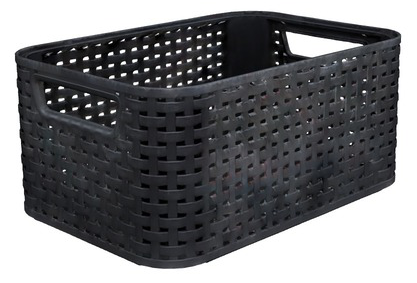}{basket}{Complex topology.}}

\revision{ 
}{%
\section{Composition of the Dataset}
The Scanned Objects dataset contains 1030 scanned objects and their associated metadata, totalling 13Gb,
licensed under the CC-BY 4.0 License \cite{CC-BY-40}.

Visual meshes are represented in Wavefront OBJ format averaging 1.4Mb (min: 0.1Mb, max: 11.1Mb) per model.
The accompanying diffuse texture maps are represented in PNG format, and average 11.2Mb (min:6.5Mb , max: 23.5Mb) per texture.

These models comprise a subset of the larger collection of models scanned by Google Cloud Robotics and 
Google Research as filtered by the pipeline process described above. Table \ref{table:model-stats} lists the breakdown of the model categories in the dataset.

}{%

\section{Properties of the Dataset}
\subsection{Dataset Composition}
The GSO dataset contains 1030 scanned objects and associated metadata, totalling 13Gb, licensed under the CC-BY 4.0 License \cite{CC-BY-40}. These models are a subset of the larger collection scanned by Google that pass our curation pipeline. Table \ref{table:model-stats} breaks down the model categories in the dataset.

Visual meshes are in Wavefront OBJ format \cite{WavefrontOBJ} averaging 1.4MB (min: 0.1MB, max: 11.1MB) per model. The accompanying diffuse texture maps are in PNG format, and average 11.2MB (min: 6.5MB , max: 23.5MB) per texture.
\subsection{Strengths of the Dataset}
Our automated pipeline enables quick generation of large numbers of models without manual processing. Because these models are scanned rather than modeled by hand, they are realistic, not idealized, reducing the difficulty of transferring learning from simulation to real \cite{ActionImageRep, RetinaGAN, RL-CycleGAN}.

Our scanner's glass platform enables scanning models from all sides including the base, unlike similar scanners with opaque platforms \cite{singh2014bigbird, kasper2012kit, lai2011large}. Similarly, models extracted from an environment \cite{janoch2013category, Dai_2017_CVPR}
will generally lack occluded regions like the base, which must be stitched.

Because our scanner reconstructs surface shape from projected patterns rather than from depth camera data, the resulting mesh has high fidelity.
Smooth surfaces are smooth and silhouette edges are accurate (\figref{object-details}) compared with meshes derived from RGB-D data which can appear mottled and irregular, particularly at silhouettes.

\subsection{Limitations of the Dataset}
Our scanner's capture area cannot accommodate objects larger than a breadbox (roughly 50 cm), so the dataset does not include larger objects, such as the chairs, automobiles, or airplanes, found in some other datasets \cite{shapenet2015, ModelBank}.
Similarly, scan resolution is limited, so very small objects cannot be modeled with reasonable fidelity. Furthermore, generated textures are diffuse: highly specular or transparent objects cannot be represented and can yield poor results.
}

\begin{table}[tbp]
\centering
\caption{Projects using Google Scanned Objects.}
{\footnotesize\begin{tabular}{|l|c|c|c|}
\hline
\textbf{Project} &\textbf{Usage} &\textbf{Renderer} &\textbf{Simulator} \\
\hline\hline
\multicolumn{4}{|l|}{\bf Vision} \\ \hline
NViSII \cite{NViSII}  & Training & NViSII & PyBullet \\
Faster R-CNN \cite{FasterRCNN}  & Training & OpenGL & N/A \\
\hline\hline
\multicolumn{4}{|l|}{\bf Graphics} \\ \hline
IBRNet \cite{IBRNet}  & Training & IBRNet & N/A \\
NeuRay \cite{NeuRay}  & Training & NeuRay & N/A \\
Diff. Stereo. \cite{goel2021differentiable} &  Validation & DS & N/A \\
\hline\hline
\multicolumn{4}{|l|}{\bf Navigation} \\ \hline
HMS \cite{HMS2020}  & Training & iGibson & iGibson \\
iGibson \cite{iGibsonChallenge2021}  & Benchmark & iGibson & iGibson \\
\hline\hline
\multicolumn{4}{|l|}{\bf Manipulation} \\ \hline
GIGA \cite{GIGA2021}  & Training & NViSII & PyBullet \\
LAX-RAY \cite{LaxRay2020} & Training & Pyrender & Trimesh \\
O2O-Afford  \cite{O2OAfford}  & Training & N/A & SAPIEN \\
\hline\hline
\multicolumn{4}{|l|}{\bf 3D Shape Processing} \\ \hline
Keypoint Deformer \cite{KeypointDeformer} & Validation & N/A & N/A \\
\hline
\end{tabular}}
\label{table:projects}
\end{table}

\section{Usage of the Dataset}
Hosting GSO on Ignition Fuel has made it possible for people inside and outside the robotics community to find, download and use the dataset in simulation and their research projects.
To our knowledge, the GSO dataset has been used in a dozen papers across ten projects spanning computer vision, computer graphics, robot manipulation, robot navigation, and 3D shape processing (Table \ref{table:projects}). 

Most projects used the GSO dataset to provide synthetic training data for learning algorithms.
The NVIDIA Scene Imaging Interface (NViSII) \cite{NViSII} and Faster R-CNN \cite{FasterRCNN} projects generated synthetic data for deep vision systems,
while the IBRNet \cite{IBRNet} and Neural Ray \cite{NeuRay} projects generated training data for deep rendering systems.

The iGibson Challenge 2021 \cite{iGibsonChallenge2021} and 
Hierarchical Mechanical Search (HMS) \cite{HMS2020} 
projects used these objects to populate scenes for robotic navigation training and benchmarking tasks.
Across many of these projects, the addition of controllable synthetic data to training pipelines either improved the overall performance of the system or improved the visual realism of the simulated environments.

Notably, three projects -
Grasp detection via Implicit Geometry and Affordance (GIGA) \cite{GIGA2021},
Lateral Access X-RAY \cite{LaxRay2020},
and Object-Object Affordance Learning \cite{O2OAfford}
- used the GSO dataset for the purpose that inspired it: generating synthetic data to train robotic grasping and manipulation tasks.
\revision{[E1, R1.1, R2.1, R2.3, R3.1, R3.2]}{
As with the vision, graphics, and manipulation tasks, the grasping data pipelines either relied on the dataset to create their scenes, or showed explicit improvements by augmenting real data.
}{These pipelines either relied on GSO to create their scenes, or showed  improvements by augmenting real data.
}

\revision{
}{%
Finally, because the dataset is scanned from real objects, the KeypointDeformer \cite{KeypointDeformer} was able to use the GSO dataset as a validation set for algorithms trained on other data sources.
}{%
An internal version of the dataset with cleaned-up transparency and specular materials 
was used by
Action Image Representation \cite{ActionImageRep}, 
RetinaGAN \cite{RetinaGAN}, 
and RL-CycleGAN \cite{RL-CycleGAN}.
The visual realism of the models, plus randomizing visual and physical properties, facilitated sim-to-real transfer learning for real-world tasks.

Finally, because the dataset is scanned from real objects, KeypointDeformer \cite{KeypointDeformer} and Differentiable Stereopsis (DS) \cite{goel2021differentiable} were able to use the GSO dataset as a validation set. DS used 50 categories in one evaluation, alongside two other evals using \cite{collins2021abo, knapitsch2017tanks}. }

\section{CONCLUSIONS}
This letter presented Google Scanned Objects,
a diverse collection of high-quality 3D scans of household objects
released in Ignition Gazebo format.
These objects have already proven useful in ten robotics and simulation projects.
We hope that the Google Scanned Objects dataset 
will be used by more robotics and simulation researchers in the future, 
and that the example set by this this dataset 
will inspire other owners of 3D model repositories
to make them available for researchers everywhere.



\section*{ACKNOWLEDGMENT}

The authors thank the Google Scanned Objects team, including Peter Anderson-Sprecher, J.J. Blumenkranz, James Bruce, Ken Conley, Katie Dektar, Charles DuHadway, Anthony Francis, Chaitanya Gharpure, Topraj Gurung, Kristy Headley, Ryan Hickman, John Isidoro, Sumit Jain, Greg Kline, Mach Kobayashi, Kai Kohlhoff, James Kuffner, Thor Lewis, Mike Licitra, Lexi Martin, Julian (Mac) Mason, Rus Maxham, Pascal Muetschard, Kannan Pashupathy, Barbara Petit, Arshan Poursohi,    Jared Russell, Matt Seegmiller, John Sheu, Joe Taylor, and Josh Weaver.
Special thanks go to James Bruce for the scanning pipeline design and Pascal Muetschard for maintaining the database of object models.


\bibliographystyle{unsrt}
\bibliography{main}  

\addtolength{\textheight}{-12cm}   


\end{document}